\documentclass[preprint,12pt,authoryear]{elsarticle}

\usepackage{amssymb}
\usepackage{amsmath}

\usepackage{graphicx, subfigure}
\usepackage[table,xcdraw]{xcolor}
\usepackage[colorlinks=true,citecolor=red, linkcolor=blue]{hyperref}
\usepackage{tabularx}
\usepackage{longtable}
\usepackage{caption} 
\usepackage[skip=0.333\baselineskip]{caption}
\usepackage{color}
\usepackage{tabularray}
\definecolor{JapaneseLaurel}{rgb}{0,0.501,0}
\usepackage{float}
\usepackage{booktabs}
\usepackage{adjustbox}
\usepackage{comment}
\usepackage{algorithm} 
\usepackage{algpseudocode} 
\usepackage[version=3]{mhchem} 
\usepackage{siunitx}
\usepackage{booktabs}
\usepackage{hyperref, url}
\definecolor{JapaneseLaurel}{rgb}{0,0.501,0}
\usepackage{tikz}
\def\checkmark{\tikz\fill[scale=0.4](0,.35) -- (.25,0) -- (1,.7) -- (.25,.15) -- cycle;} 
\newcommand{\tikzxmark}{%
\tikz[scale=0.23] {
    \draw[line width=0.7,line cap=round] (0,0) to [bend left=6] (1,1);
    \draw[line width=0.7,line cap=round] (0.2,0.95) to [bend right=3] (0.8,0.05);
}}

\newcommand{\mcrot}[4]{\multicolumn{#1}{#2}{\rlap{\rotatebox{#3}{#4}~}}} 

\journal{Image and Vision Computing}

\begin{document}

\begin{frontmatter}



\title{Designing and Generating Diverse, Equitable Face Image Datasets for Face Verification Tasks} 

\author{Georgia Baltsou} 
\author{Ioannis Sarridis}
\author{Christos Koutlis}
\author{Symeon Papadopoulos}

\affiliation{organization={Information Technologies Institute, CERTH},
            city={Thessaloniki}, 
            country={Greece}}

\begin{abstract}
Face verification is a significant component of identity authentication in various applications including online banking and secure access to personal devices. The majority of the existing face image datasets often suffer from notable biases related to race, gender, and other demographic characteristics, limiting the effectiveness and fairness of face verification systems. In response to these challenges, we propose a comprehensive methodology that integrates advanced generative models to create varied and diverse high-quality synthetic face images. This methodology emphasizes the representation of a diverse range of facial traits, ensuring adherence to characteristics permissible in identity card photographs. Furthermore, we introduce the Diverse and Inclusive Faces for Verification (DIF-V) dataset, comprising 27,780 images of 926 unique identities, designed as a benchmark for future research in face verification. Our analysis reveals that existing verification models exhibit biases toward certain genders and races, and notably, applying identity style modifications negatively impacts model performance. By tackling the inherent inequities in existing datasets, this work not only enriches the discussion on diversity and ethics in artificial intelligence but also lays the foundation for developing more inclusive and reliable face verification technologies.
\end{abstract}

\begin{graphicalabstract}
\includegraphics[width=16cm]{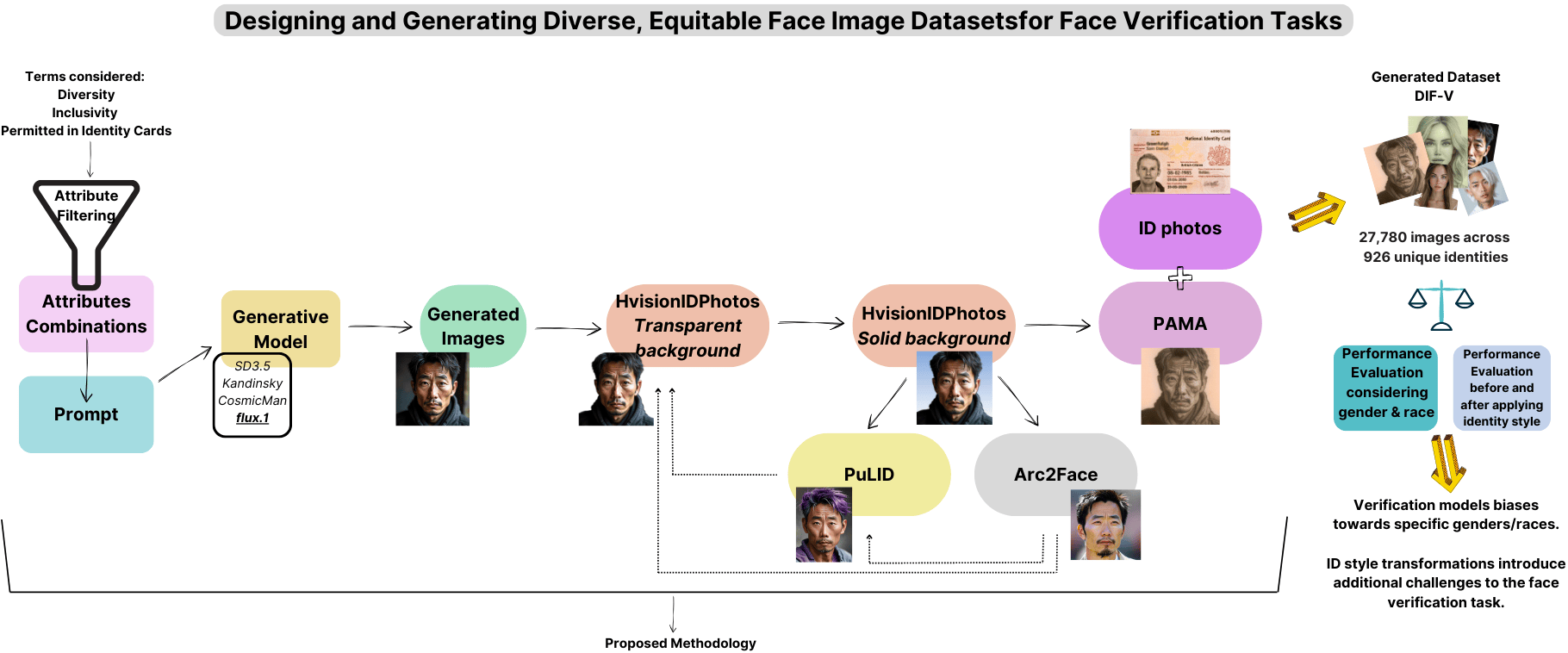}
\end{graphicalabstract}

\begin{highlights}
\item Proposal of a methodology that combines advanced generative models to produce high-quality synthetic images representing a broader spectrum of facial diversity, considering essential characteristics allowed in identity card photos. 
\item Introduction of the Diverse and Inclusive Faces for Verification (DIF-V) dataset, comprising 27,780 images across 926 unique identities, thus serving as a benchmark for future face verification research.
\item Our analysis reveals that existing verification models exhibit biases toward certain genders and races, and notably, applying identity style transformations negatively impacts model performance.
\end{highlights}

\begin{keyword}
synthetic dataset \sep diversity \sep AI \sep face verification

\end{keyword}

\end{frontmatter}


\section{Introduction}
\label{sec:intro}
Face verification constitutes a task of face image analysis that involves comparing a person’s current facial characteristics with their stored biometric template to verify their identity \citep{thom2023doppelver}. This task is critical in situations where verifying an individual's identity is required, for instance, in applications like online banking \citep{arunadevi2023online}, unlocking personal devices \citep{venkata2021intelligent} or document verification (verify whether passport and driver's license photos belong to the same person) \citep{paliwal2020faceid}. In order to build accurate face verification systems, appropriate face image datasets should be available for training and testing. 

Existing face image datasets used as benchmarks for face verification, are typically collections of real-world images, often sourced from the Internet. Such real-world datasets present significant challenges. First of all, the majority of these images portray famous people with make-up who also are light enhanced, among others \citep{kortylewski2019analyzing, melzi2024synthetic}. As a result, the datasets are biased towards attributes such as race, gender, and age, or to specific photo properties (e.g., scene composition, lighting) and do not accurately reflect the distribution of face photos in the real world \citep{sarridis2023towards}. Furthermore, since such datasets mostly depict famous people, they introduce two more significant concerns apart from bias. First, their reliance on famous people results in a demographic imbalance, skewing the data towards specific genders and races \citep{thabtah2020data}. Second, as these images are commonly sourced from the Internet, they are often used without a proper license \citep{colbois2021use}.

\begin{figure}[ht]
\centering
\subfigure[][]{
\label{fig:inf-a}
\includegraphics[height=1.2in]{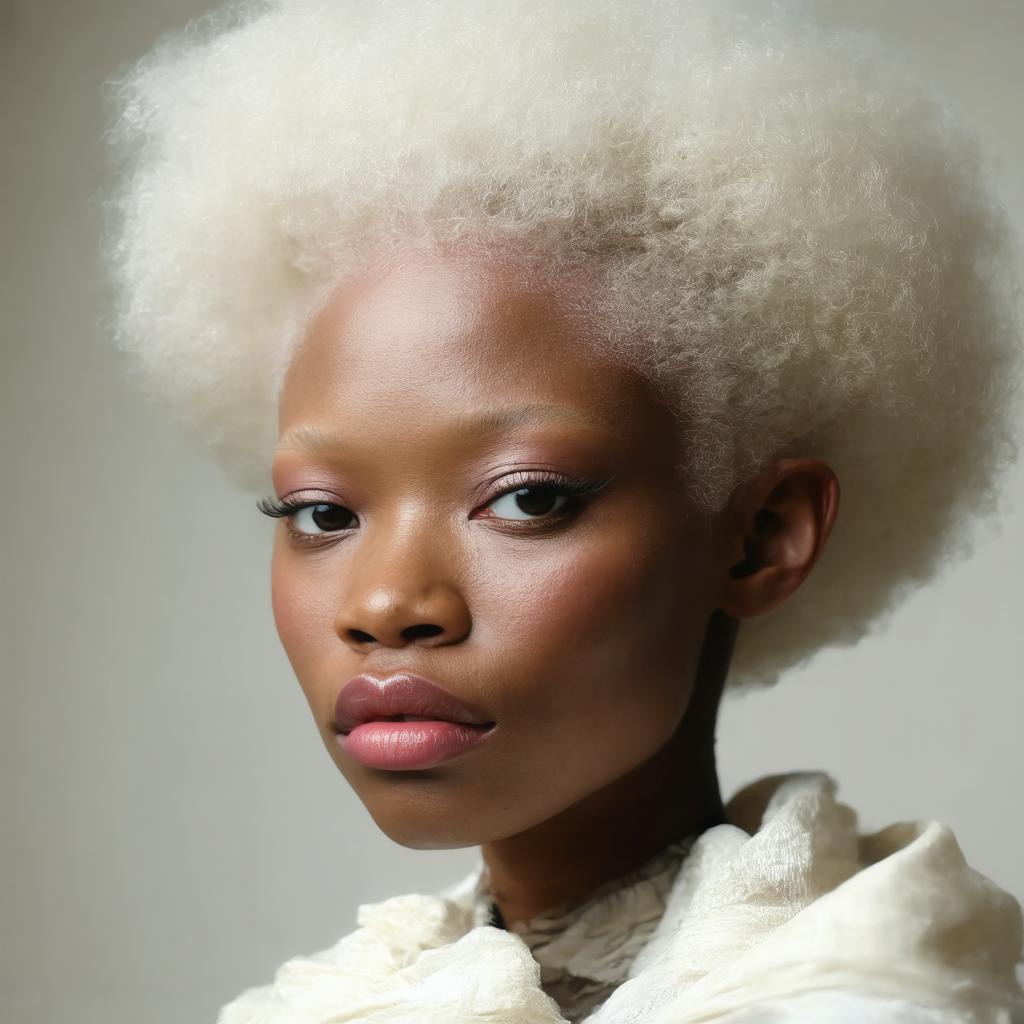}}
\subfigure[][]{
\label{fig:inf-b}
\includegraphics[height=1.2in]{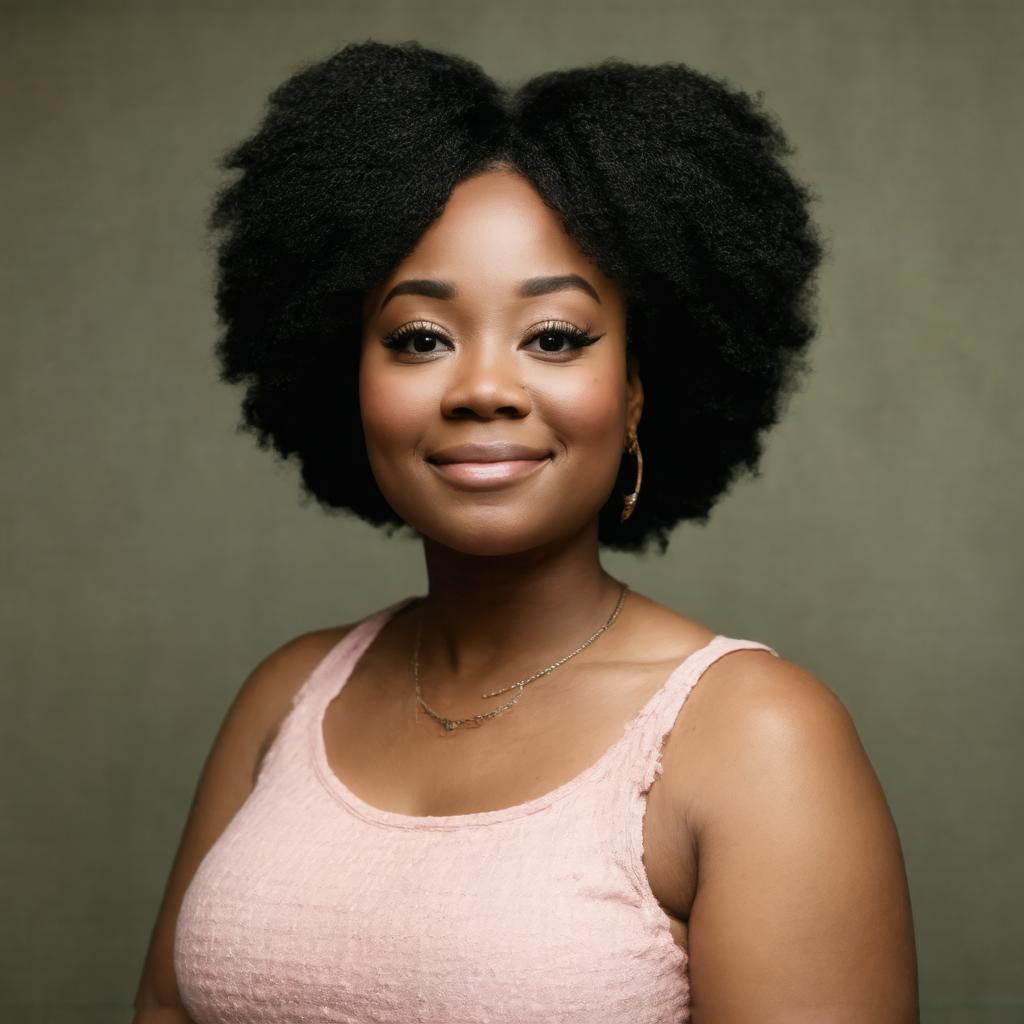}}
\subfigure[][]{
\label{fig:inf-c}
\includegraphics[height=1.2in]{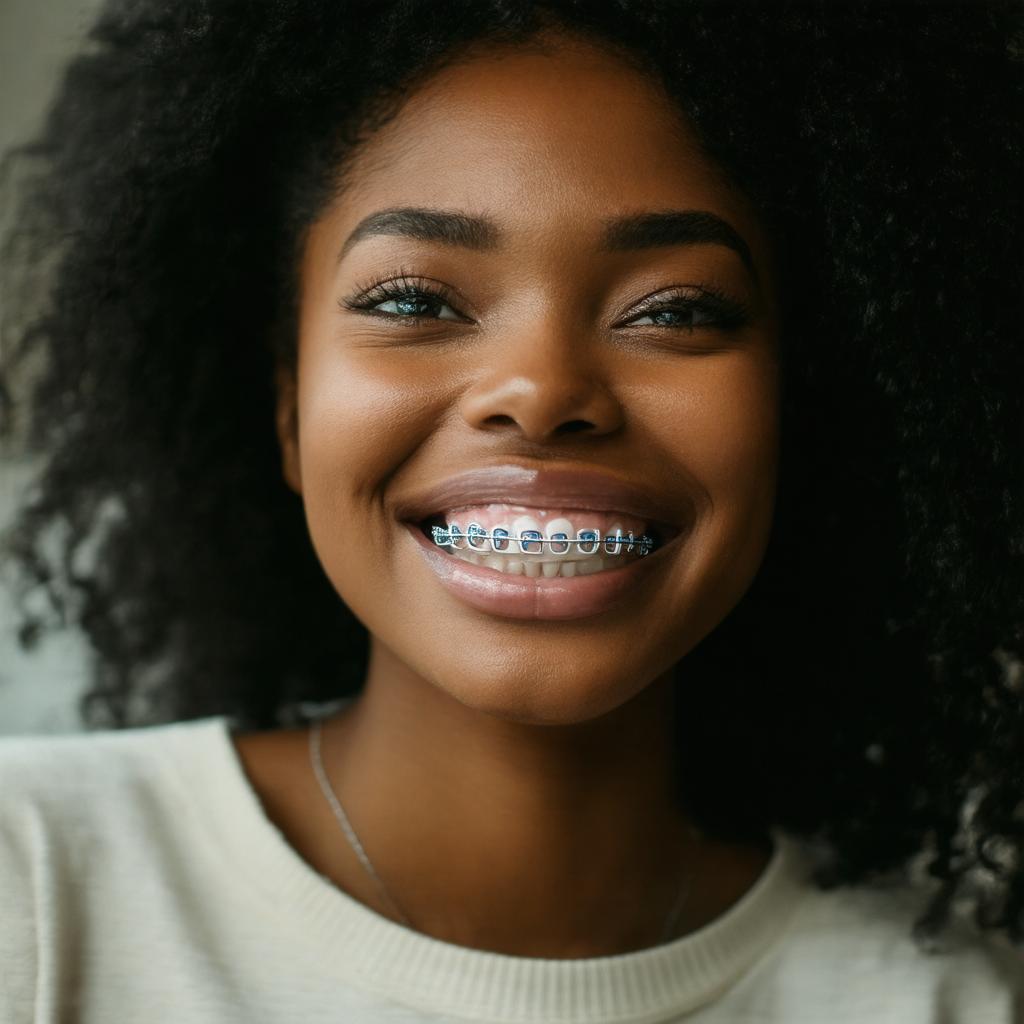}}
\subfigure[][]{
\label{fig:inf-d}
\includegraphics[height=1.2in]{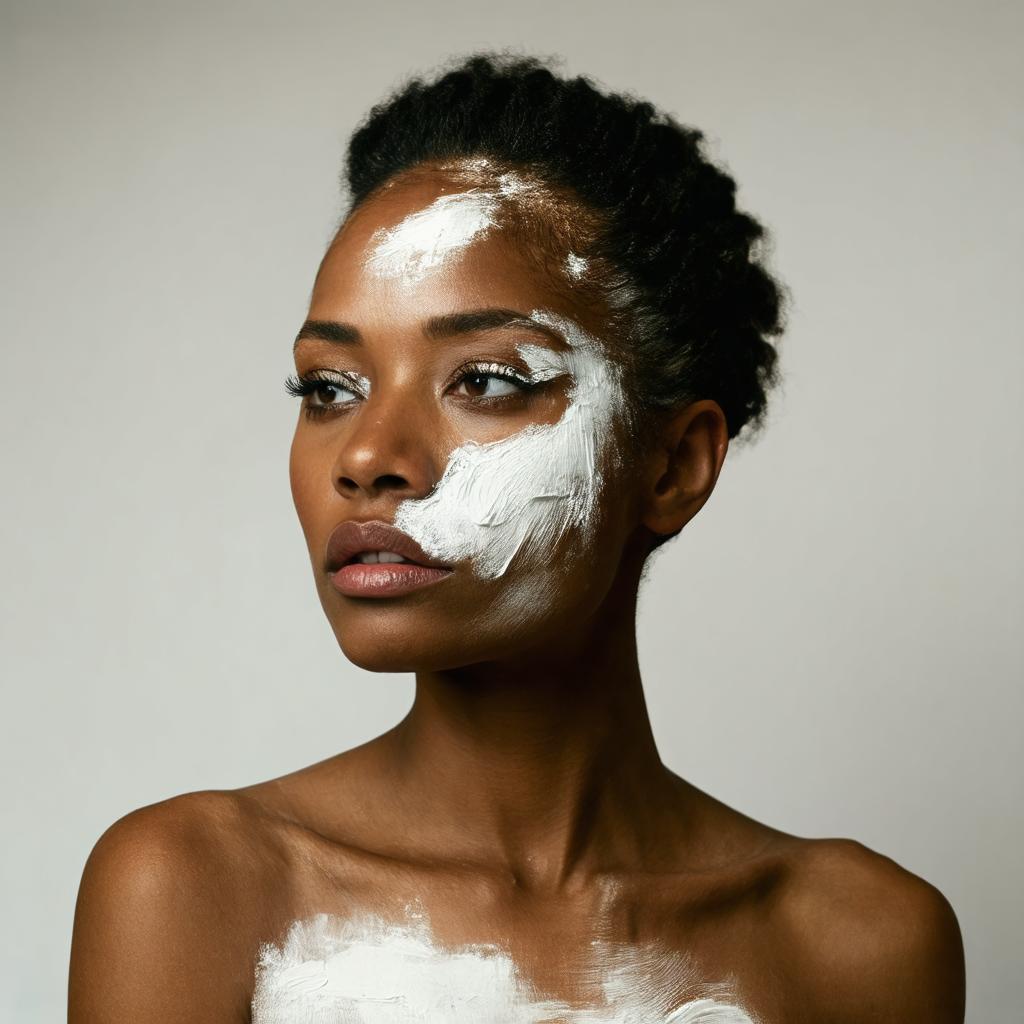}}
\caption[]{Example generated images with T2I model Stable Diffusion 3.5 and the less common features of:
\subref{fig:inf-a} albinism,
\subref{fig:inf-b} down syndrome,
\subref{fig:inf-c} tooth braces and
\subref{fig:inf-d} vitiligo. Prompt adherence remains a key challenge.}
\label{fig:non_common}
\end{figure}

Synthetic face images, on the other hand, have the potential to alleviate the above issues. Specifically, because image generation is controlled, the number of images per category can be balanced and reflect a variety of target attributes that represent the variation of the face in the real world \citep{serna2021insidebias,terhorst2021comprehensive,melzi2024synthetic}. Furthermore, AI-generated face images nowadays are highly realistic and difficult to tell from real face photos \citep{miller2023ai}. Previous works have already demonstrated the potential of synthetic face images for face recognition and verification tasks \citep{boutros2023synthetic, melzi2024frcsyn,  melzi2024frcsyn_b}. However, work focusing on the generation of synthetic face images \citep{deng2020disentangled, colbois2021use, wood2021fake, qiu2021synface, grimmer2021generation, mekonnen2023balanced, bae2023digiface, kim2023dcface} also faces some challenges. Although most synthetic images today are highly realistic, some may exhibit imperfections or deviations from natural images \citep{hao2024synthetic}. In addition, synthetic images are generated by tools which were trained on images retrieved from the Internet. As a result, generative models bring their own biases \citep{zhou2024bias, hao2024synthetic}. Furthermore, the majority of synthetic datasets consider diversity as the inclusion of faces from people of different age, gender, or skin tone and not in terms of other biometrics, and non-permanent traits \citep{karkkainen2021fairface, grimmer2021generation}. Besides, adherence to input prompts, especially for less common features, remains a key challenge for Text-to-Image (T2I) models. Figure~\ref{fig:non_common} provides examples of this issue. Finally, identity preservation, the ability to generate novel images of a specific individual from reference images while adhering to textual prompts, remains a significant challenge for T2I models \citep{chen2024id, li2024photomaker}.

In the present work, we provide an approach for producing equitable and realistic synthetic face image datasets that cover a broad spectrum of natural facial variety and may be used in face verification tasks. This is accomplished by combining multiple existing models into a methodology that produces realistic and diversified facial images. Since the generated images are designed to be utilized in face verification tasks, diversity is limited to characteristics that are acceptable in identity card photos, namely race, gender, weight, hair color and style, eye color, mustache/beard, tattoo and other permanent traits like wrinkles, freckles or (burn/acne) scars.
This research builds on our previous work \cite{10581864}, where we introduced a methodology for generating diverse synthetic face image datasets using Stable Diffusion 2.1 \citep{Rombach_2022_CVPR}. That work resulted in the Stable Diffusion Faces Dataset (SDFD), which comprises $1,000$ images that showcase a wide range of facial diversity in terms of race, gender, age, accessories, makeup and emotions. However, SDFD was limited in size and image generation methods, producing images of distinct identities and thus not suitable for face verification. The current work addresses these limitations by focusing on identity preservation and creating a dataset specifically designed for verification tasks, while retaining the diversity achieved in our previous efforts. We incorporate four generative models to produce face images, two models for facial attribute manipulation, and two additional models for identity card photo formatting.

Besides, following the suggested methodology, we provide a new face image dataset that can be used as a benchmark set in face verification tasks. More precisely, we created \textit{DIF-V} (Diverse and Inclusive Faces for Verification) dataset. This dataset contains $27,780$ different face images of $926$ unique identities, showcasing individuals of varying races, genders, weights, and other permanent characteristics.

The contributions of this work are threefold:
\begin{itemize}
    \item From a \textbf{methodology perspective}, we propose a framework for designing and generating diverse and demographically balanced face image datasets. This ensures equitable representation across different demographic groups, fostering fairness in face verification research.
    \item From a \textbf{practical perspective}, we introduce the DIF-V dataset, which consists of $27,780$ images spanning $926$ unique identities. DIF-V serves as a novel benchmark dataset specifically designed to mitigate bias and promote inclusivity in future face verification systems.
    \item From an \textbf{experimental perspective}, we conduct a comprehensive evaluation of state-of-the-art face verification models (ArcFace and Facenet) in the DIF-V dataset. Our results reveal significant performance disparities across demographic groups. In addition, style transformations to face images degrade the performance of both models, indicating increased challenges for these models in verification tasks.
\end{itemize}

\section{Related Work}
\label{sec:related}

\subsection{Face Image Datasets for Face Verification}

The literature contains a large number of datasets that are used as benchmarks for face verification tasks. We decided to list the most popular ones here, all of which happen to comprise real face images. 
LFW \citep{huang2008labeled} constitutes a face image dataset that covers the range of conditions typical of everyday life. It contains $14,089$ Internet images that correspond to $5,749$ unique identities. However, LFW predominantly consists of white male celebrities.
RFW \citep{wang2019racial} is a real face image dataset that includes approximately $40,000$ images of almost $12,000$ unique identities. RFW is a bias-aware benchmark; nevertheless, it uses broad labels of race categories. For example Asian race mixes East and Southeast Asians. It also lacks intersectional diversity.
Another face image dataset that focuses on the precise age of the depicted face is ageDB \citep{moschoglou2017agedb}. It is derived from the Internet and includes $16,488$ images of $568$ distinct identities. Like LFW, ageDB consists mostly of famous individuals and, like RFW, it also lacks intersectional diversity.
CFP-FP \citep{sengupta2016frontal} includes balanced frontal and profile real face images of famous people, sourced from the Internet. Includes $7,000$ images of $350$ unique identities. However, it does not include diverse demographics since it consists mainly of Caucasian and Asian celebrities.
ROF \citep{erakiotan2021recognizing} contains face image samples with real life upper- and lower-face occlusions (i.e. face masks and sunglasses). Includes $5,559$ images of $180$ identities. Nonetheless, it mostly includes images of Caucasian and Asian people.
DoppelVer \citep{thom2023doppelver} contains $27,967$ images comprised of look-alike individuals derived from the Internet and also suffers from celebrity bias.

\subsection{Text-to-Image Generation Methods}

Text-to-image (T2I) generation is rapidly advancing, as seen by the rapid growth of T2I published research \citep{yang2023diffusion, alhabeeb2024text}. Most existing approaches rely on foundation models designed for general-purpose domains. Among these, diffusion methods have become very popular due to the quality and diversity of generated images \citep{Rombach_2022_CVPR, razzhigaev2023kandinsky, yang2023diffusion, lin2024diffusion, chen2025comprehensive}. 
Furthermore, diffusion models have been empirically found to have higher text-image alignment values compared to other types of models like Generative Adversarial Networks (GANs) \citep{kang2023scaling, lee2023holistic}.

The complexity of human anatomy and appearance motivated the creation of text-driven human image generation models. There are works focusing on generating humans or more specifically faces. Many of these models, focus on the fashion domain \citep{jiang2022text2human, lin2023fashiontex}. Concerning face image generation, some of the first methods were based on GANs \citep{gecer2018semi, shen2018facefeat, deng2020disentangled} and focused on the control of desired face properties via learning disentangled representations in the latent space.  Diffusion based methods for synthetic face image generation include methods such as \citep{kim2023dcface} and CosmicMan \citep{li2024cosmicman}, which is a human-centric foundation model. 

\subsection{Image Editing Methods}

Image editing constitutes another interesting and popular class of generative models. In contrast to T2I generation which produces new images given some text input, image editing involves changing an image's appearance or content. There are many fields where image editing is essential, such as advertising or scientific research. 

Personalized generation methods, or methods that generate images based on an original image that portrays the same identity as the original, are a subset of image editing techniques. The majority of methods of this category require an image as input and some text outlining the changes that must be made to the original image.
Indicative methods of this category are IP-Adapter \citep{ye2023ip}, InstantID \citep{wang2024instantid} that uses the ArcFace \citep{deng2022arcface} face recognition model to ensure ID fidelity. Similar works are PhotoVerse \citep{chen2023photoverse}, and PhotoMaker \citep{li2024photomaker}, which are based on diffusion models for tuning-free personalization. The PuLID method \citep{guo2024pulid} also relies on ArcFace \citep{deng2022arcface} along with SDXL \citep{podellsdxl} to achieve better ID fidelity and editability compared to previous methods. 
In addition, there are methods which are based only on face embeddings retrieved from a face recognition model to generate ID variations. Arc2Face \citep{papantoniou2024arc2face} is such a method.

Another subset of image editing techniques is that of style transfer. Style transfer is an already well-examined image processing technique that aims to apply the artistic style of a reference image to a target image while preserving the content of the target image \citep{ding2024regional}. AdaIN \citep{huang2017arbitrary} is a method that adaptively applies the mean and standard deviation of each style feature to share the same distribution. The authors of \cite{park2019arbitrary} proposed the Style-Attentional Network (SANet) to match the content and style features. AdaAttN \citep{liu2021adaattn} method managed a better trade-off between style pattern transferring and content structure preservation. StyleFormer \citep{wu2021styleformer} constitutes a patch-based method that uses feature-level information to transfer styles with transformer-based encoding and decoding mechanisms. Progressive attentional manifold alignment (PAMA) \citep{luo2022progressive} repeatedly applies attention operations to dynamically rearrange style features according to the spatial distribution of the content. ST2SI \citep{li2024st2si} is a recent style transfer method that uses Spatial Interactive Convolution (SIC) and Spatial Unit Attention (SUA) to improve the representation of both content and style. Finally, the framework proposed by \cite{moussa2025face} transforms selfie images into document-style ones by combining style banks of document templates and face-swapping generative models. 

There are also techniques aimed at deepfake generation or modification. Such methods typically generate highly realistic images, often involving existing human faces,  raising serious ethical issues \citep{pei2024deepfake}. Most of these approaches fall into one of the following categories: \textit{i. face swapping} and \textit{ii. facial attribute editing}. Face swapping constitutes the process of identity exchange between two faces. A recent work in this category is StableSwap \citep{zhu2024stableswap}, which is based on GAN technology. However, it was found that it cannot handle extreme differences in skin color in the face \citep{pei2024deepfake}. DiffFace \citep{kim2025diffface} is another such method, but based on a diffusion model. Nonetheless, it was found to not keep the face lighting stable \citep{pei2024deepfake}. Facial attribute editing methods focus on modifying particular facial traits of an image. Recent representative works of this category are \cite{huang2023collaborative} and SDGAN \citep{huang2024sdgan}. The GAN-based method of SDGAN manages to achieve good attribute preservation and precise modification but diffusion methods like \cite{huang2023collaborative} achieve almost excellent results.

\section{Methodology for Generating a Synthetic Face Image Dataset}
\label{sec:process}
In the present section, we present the proposed methodology to generate diverse and equitable face image datasets suitable for face verification. 

\subsection{Generating Images from Text Input}
 
In the first step, a set of initial face images need to be generated utilizing a T2I model $\mathcal{M}$. 

Specifically, the first step involves identifying and selecting the attributes $\mathcal{A}$ that are most relevant to our problem. At this step, a list of $\mathcal{A}$ options that someone wants to be depicted in the final face images is formed. $\mathcal{A}$ options should be carefully selected and then filtered to eliminate specific words or phrases.
In the present work, since the focus is on face verification tasks, we selected $\mathcal{A}$ to include permitted features in identity card photos. $\mathcal{A}$ used to produce facial images are listed in Table~\ref{tab:prompts}. In particular, the race options were selected primarily based on the work of \cite{karkkainen2021fairface}. Additionally, we incorporated \textit{Pacific Islanders} as a distinct racial group due to their unique facial characteristics. For the remaining attributes, we empirically determined the final options through an iterative process, using previous research work as the basis \citep{samangouei2017facial, luccioni2024stable, microsoft, terhorst2021comprehensive}. Specifically, we tested multiple textual prompts and retained only those that reliably generated images that matched the intended attribute. For example, in the case of non-binary gender, we evaluated $20$ different words or phrases  before selecting the four options listed in Table~\ref{tab:prompts}.
The attribute ``age'' is implicitly encoded through correlated facial characteristics rather than explicitly specified. For example, gender descriptors, such as baby, boy, and girl, inherently constrain the age range. Similarly, attributes such as white hair and wrinkles serve as indirect but reliable indicators of advanced age. 


\begin{table}[H]
\small
  \begin{adjustbox}{center,max width=\linewidth}
    \begin{tabular}{p{2.6cm} p{10cm} }
      \toprule
      \bf  Attribute $\mathcal{A}$  & \bf  Options\\
      \midrule                                                                                                                                               
gender           & androgynous person, baby, boy, gender fluid person, girl, man, person having an androgynous appeal, person having male and female characteristics, woman \\ \hline

\rowcolor{lightgray}race             & Black, East Asian, Indian,  Latino, Middle Eastern, Pacific Islander, Southeast Asian, White                                                             \\ \hline

vision           & color contact lenses, glasses                                                                                                                                                         \\ \hline
\rowcolor{lightgray}eyes color           & green, blue, brown, black                                                                                                                                                          \\ \hline
hair color      & auburn, bald, black, blonde, blue, brown, green, pink, purple, red, white                                                                                                                                                         \\ \hline
\rowcolor{lightgray}hair style       &  curly, dreadlocks, long, short, straight                                                                                                                                                         \\ \hline
facial hair      & beard, moustache                                                                                                                                                         \\ \hline
\rowcolor{lightgray}other facial     & face tattoo                                                                                                                                                         \\ \hline
accident      & acne scars, burn scars, skin with freckles and imperfections, wrinkles                                                                                                                                                         \\ \hline

\rowcolor{lightgray} weight & normal weight, slightly chubby, over-weighted                                                                                                                                                         \\ \hline
analysis         & 4K, 8K, Ultra HD                                                                                                                                                         \\ \hline
\rowcolor{lightgray}camera           & Canon Eos 5D, Fujifilm XT3, Nikon Z9, shot on iPhone                                                                                                                                                         \\ 

      \bottomrule
\end{tabular}
\end{adjustbox}        
\caption{Terms used in prompts. Options are presented in alphabetical order.}\label{tab:prompts}                        
\end{table}


$\mathcal{A}$ options are then combined to form the input prompt of the T2I $\mathcal{M}$. $\mathcal{A}$ options were combined to maximize facial diversity while maintaining realism. Although the combinations appear to be largely randomized, certain restrictions were applied to ensure attribute compatibility. For instance, hair style options could be combined with a hair color attribute option, except for bald. 

In addition to the $\mathcal{A}$ options, some other terms were used in all experiments to improve the realism of the generated images. These terms, hereafter called universal, are the following: \textit{textured skin, remarkable detailed pupils, realistic dull skin noise, visible skin detail, skin fuzz, dry skin, tone mapping, one-color background}.
The input prompt can also include negative prompts to prevent the inclusion of unwanted traits in the final image.

All T2I generative $\mathcal{M}$ employed are diffusion-based.
Namely, we use Stable Diffusion version $3.5$ \citep{sd35}, CosmicMan \citep{li2024cosmicman}, Kandinsky \citep{razzhigaev2023kandinsky}, and FLUX.1 \citep{flux}. We selected these $\mathcal{M}$ over other competing models from the literature, including \cite{Tumanyan_2023_CVPR, wang2024characterfactory, esser2024scaling} due to their improved fidelity and higher realism. 

Generally, diffusion models work by breaking down the image generation process into various denoising steps. The process involves adding random noise to an image (forward process), and gradually removing it in denoising steps to generate a final image (backward process). During training, the neural network learns to predict and remove the noise added in each forward step, ultimately reconstructing the original image from its noisy state. This bidirectional framework enables the generation of high-quality images by progressively refining random noise into coherent outputs.

In the $\mathcal{M}$ chosen, we experimentally selected two parameters: the number of inference steps $i$ and the classifier-free guidance scale (CFG weight). Concerning $i$, typically the more steps, the better the result, but the longer the creation takes. In particular, the $i$ value controls how many times the model iteratively refines the image during denoising. The CFG weight affects the generator model's amount of flexibility when generating images, since the higher the weight, the greater the level of control by the provided prompt. In particular, CFG weight controls how strongly the model adheres to the text prompt vs. allowing for creative freedom. The parameter values vary depending on the model used. Thus, the number of $i$ and the guidance scale (CFG weight) were selected based on empirical tests and in accordance with the recommendations provided by the authors of $\mathcal{M}$'.  Table~\ref{tab:parameters} shows the corresponding values that provided the best trade-off between prompt adherence and visual quality.


\begin{table}[H]
  \centering
  \begin{tabular}{ p{2.8cm}  p{2.5cm}  p{2.2cm}   }
    \toprule
    \bf  Generative Model $\mathcal{M}$  & \bf  Inference steps $i$ & \bf CFG weight  \\
    \midrule

SD3.5        & $28$  & $3.5$ \\ 
Kandinsky    & $75$  & $4$   \\ 
CosmicMan    & $30$  & $7.5$    \\ 
Flux.1       & $50$  & $3.5$   \\ 

\bottomrule
  \end{tabular}
\caption{Values of $i$ and CFG weight used depending on the model $\mathcal{M}$. }
  \label{tab:parameters}
\end{table}


The generation process often must be repeated numerous times before a user achieves a desired result. The number of repetitions required depends on the specific Text-to-Image (T2I) model, the prompt's context, and the target application. Some T2I models exhibit superior prompt adherence, yielding satisfactory images in fewer attempts. Besides, the application's requirements also influence repetitions needed; for instance, generating frontal face portraits (as used in identity cards) is typically easier for models than generating profile views. In this work, we repeated the generation process an average of 20 times per image. Each resulting image was first automatically filtered using a quality score \textcolor{red}{(to be discussed further)} and then manually verified to eliminate failures, such as images depicting multiple faces instead of a single one.

\begin{figure}
\centering
\subfigure[][]{
\label{fig:ex-a}
\includegraphics[height=0.9in]{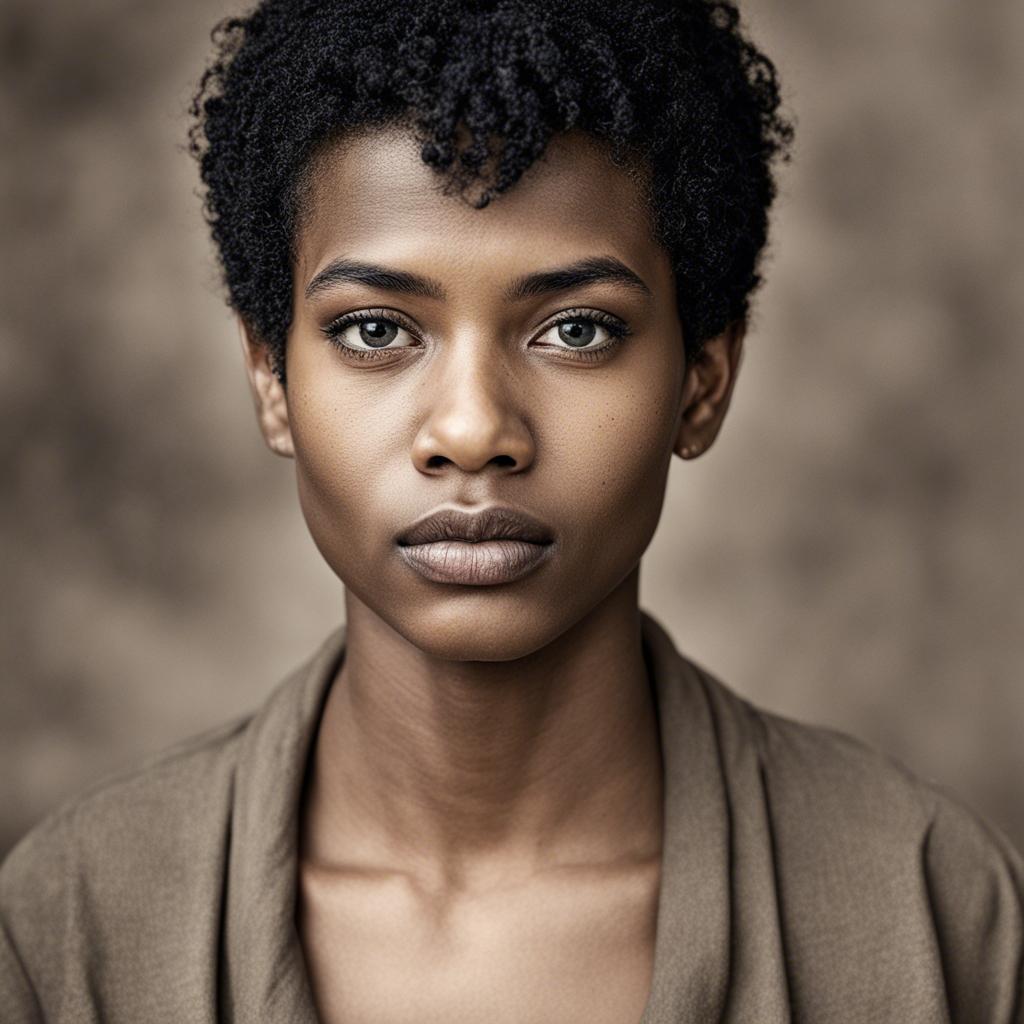}}
\subfigure[][]{
\label{fig:ex-b}
\includegraphics[height=0.9in]{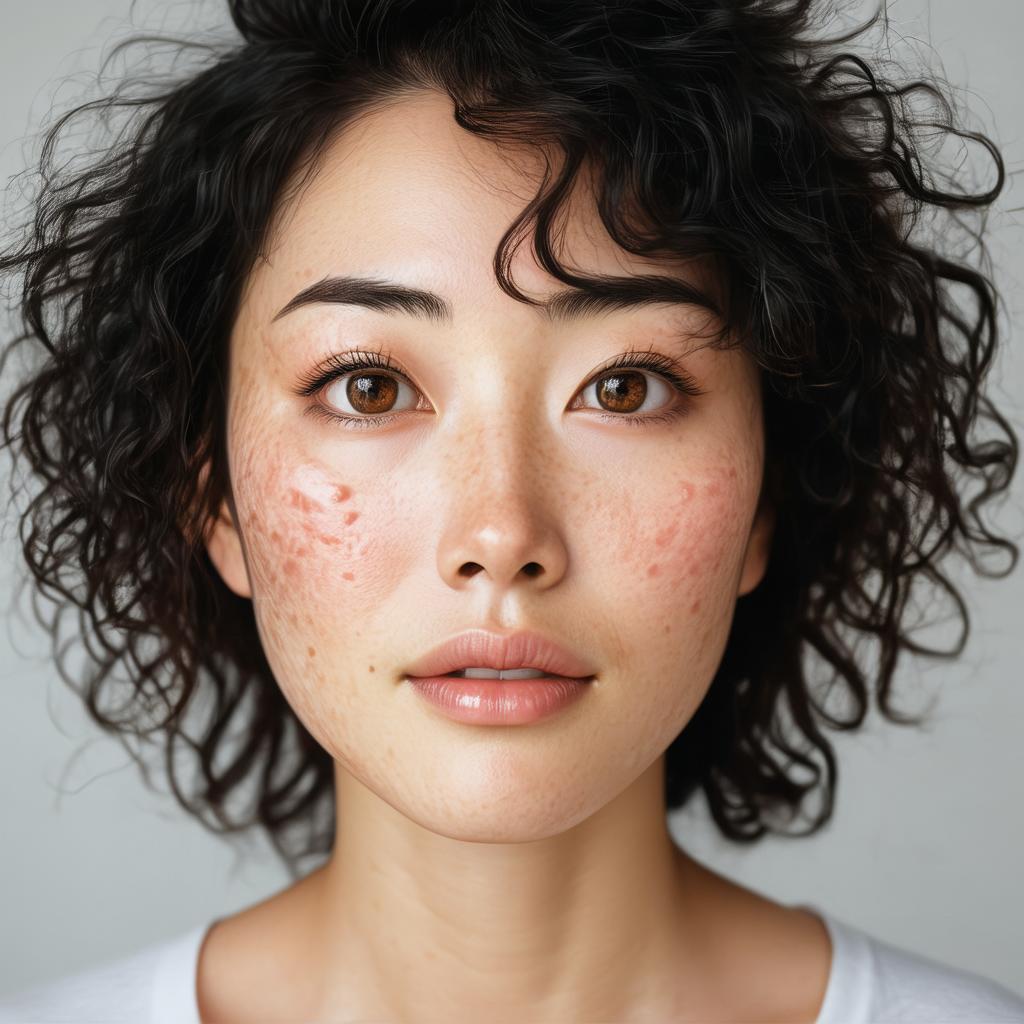}}
\subfigure[][]{
\label{fig:ex-c}
\includegraphics[height=0.9in]{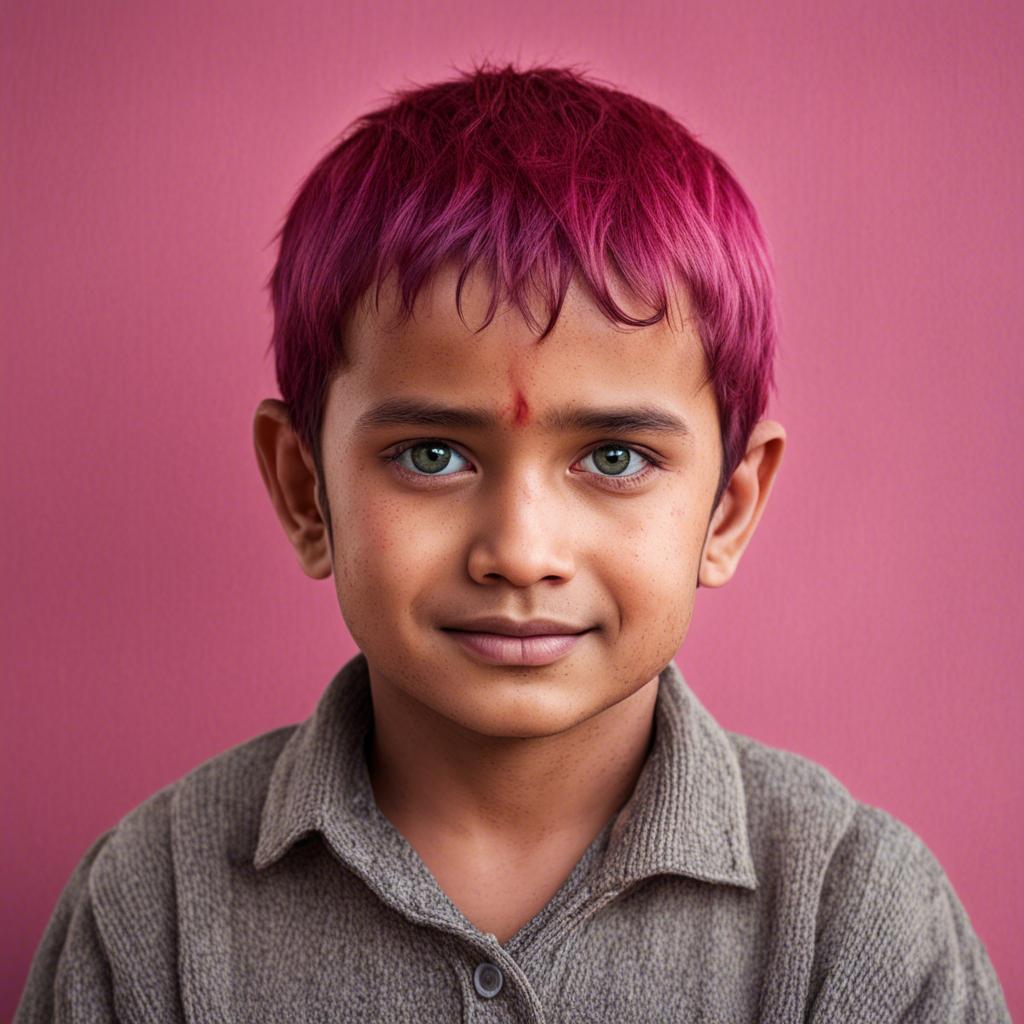}}
\subfigure[][]{
\label{fig:ex-d}
\includegraphics[height=0.9in]{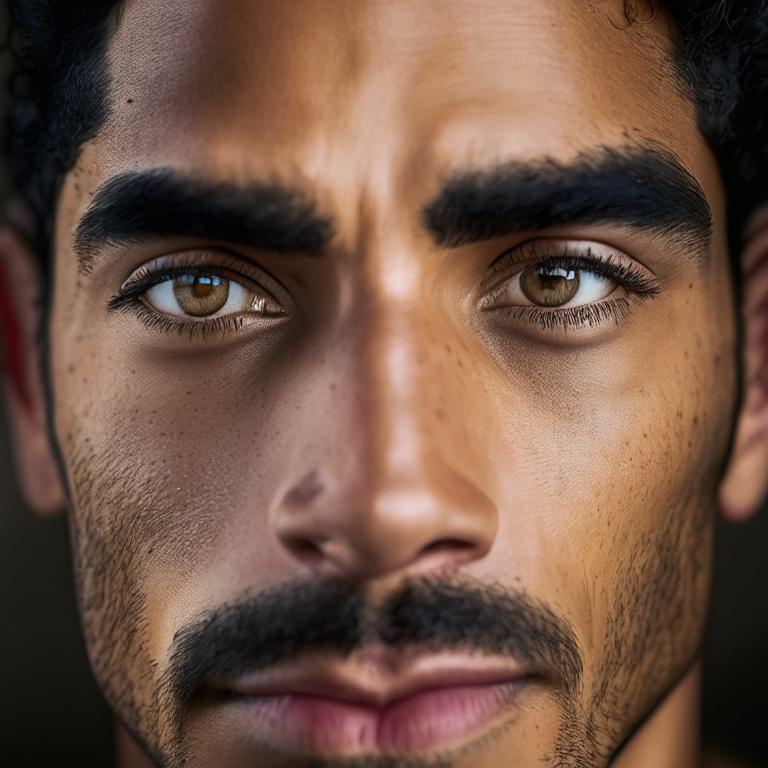}}
\subfigure[][]{
\label{fig:ex-e}
\includegraphics[height=0.9in]{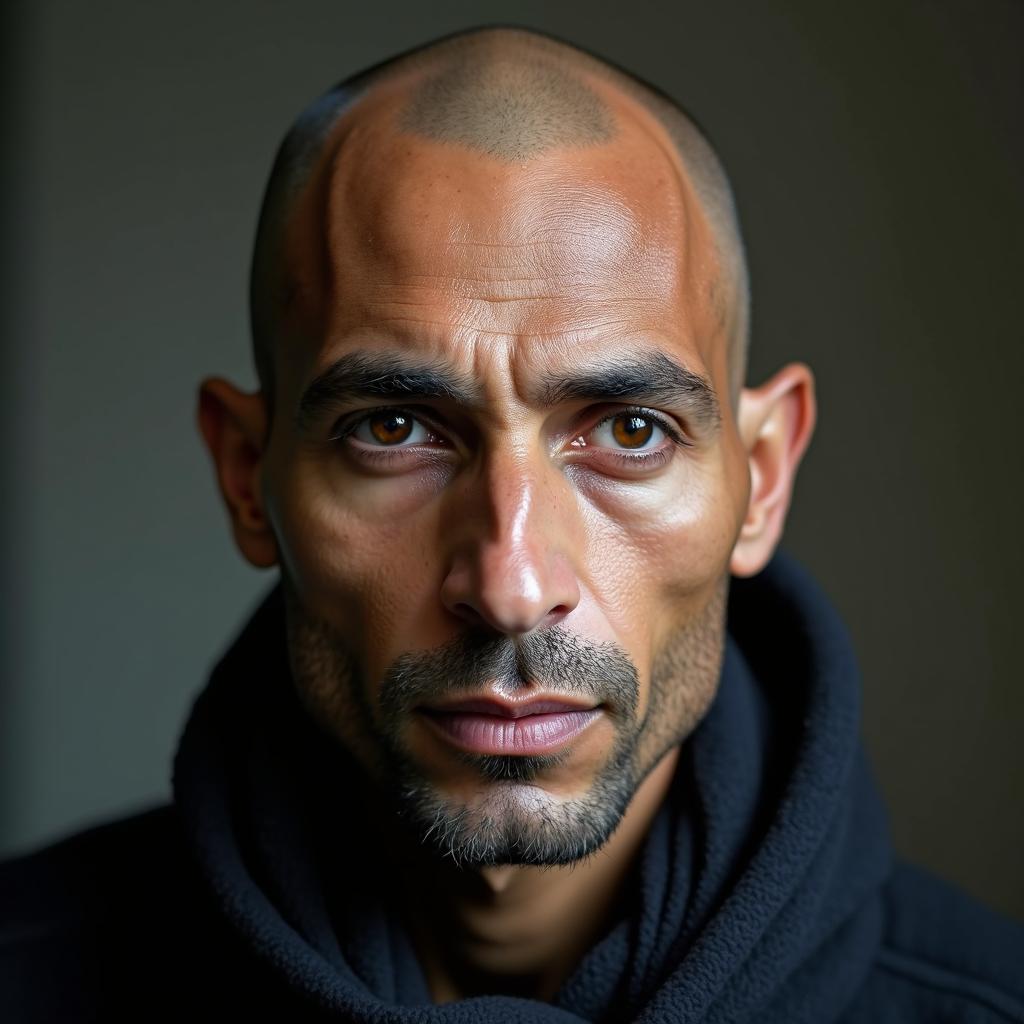}}
\subfigure[][]{
\label{fig:ex-f}
\includegraphics[height=0.9in]{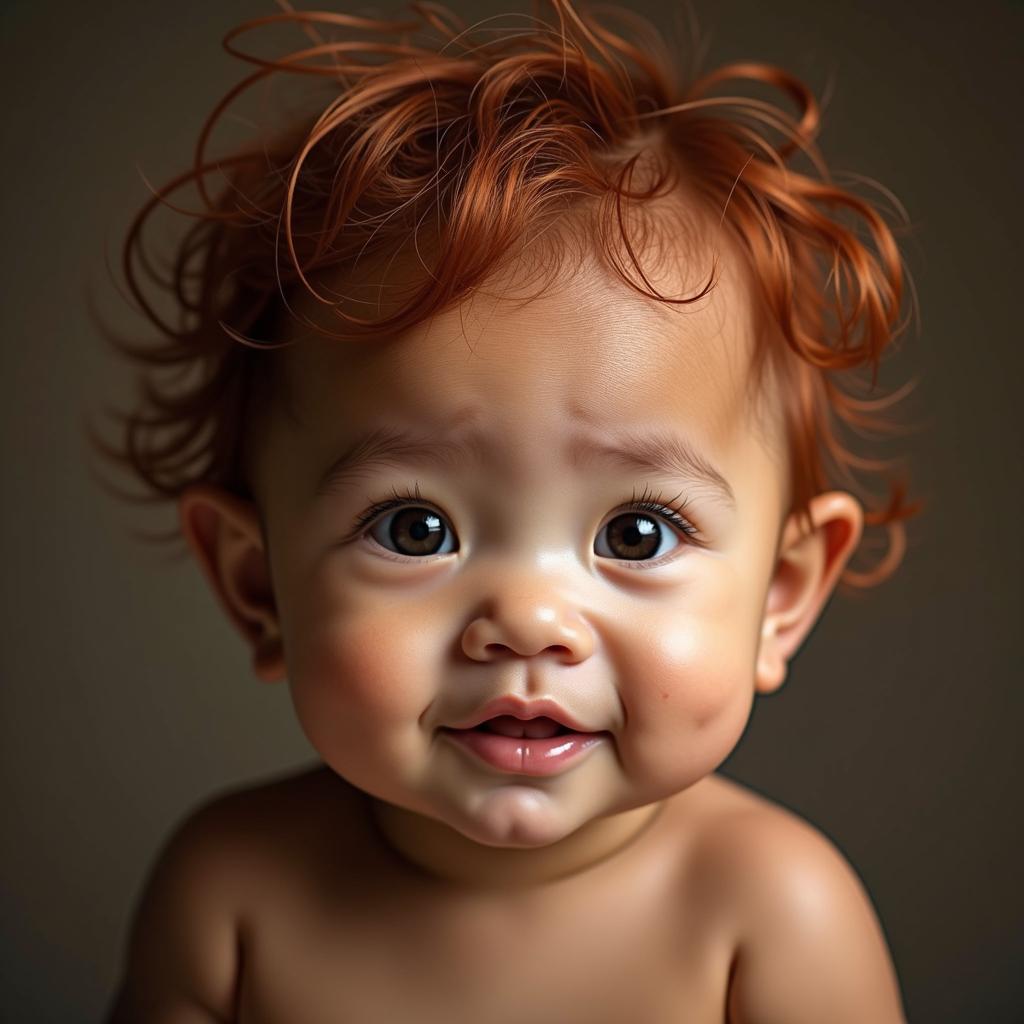}}
\subfigure[][]{
\label{fig:ex-g}
\includegraphics[height=0.9in]{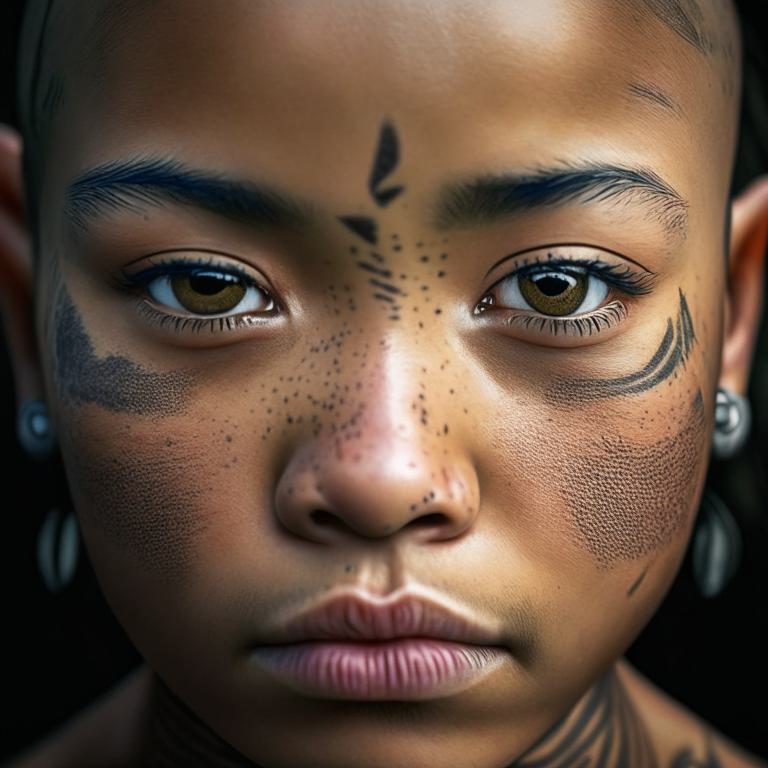}}
\subfigure[][]{
\label{fig:ex-h}
\includegraphics[height=0.9in]{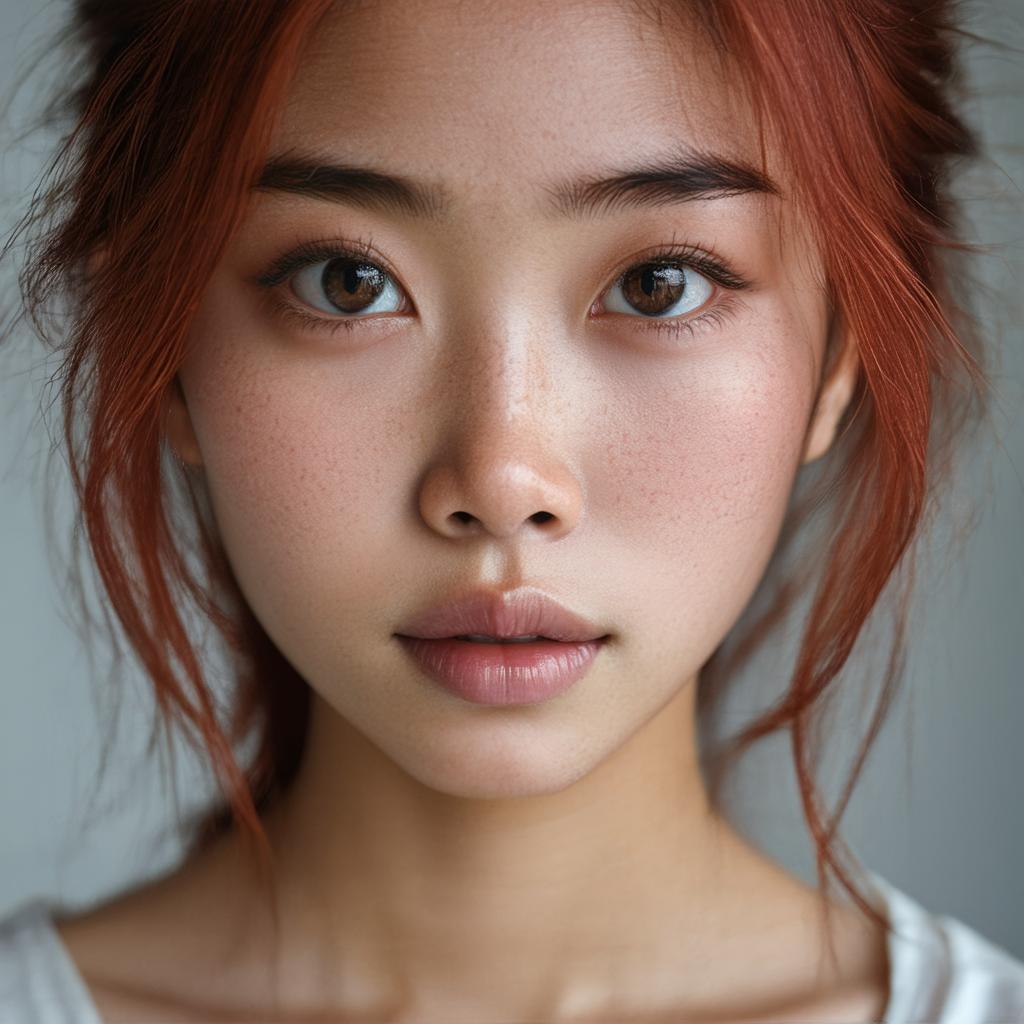}}
\subfigure[][]{
\label{fig:ex-i}
\includegraphics[height=0.9in]{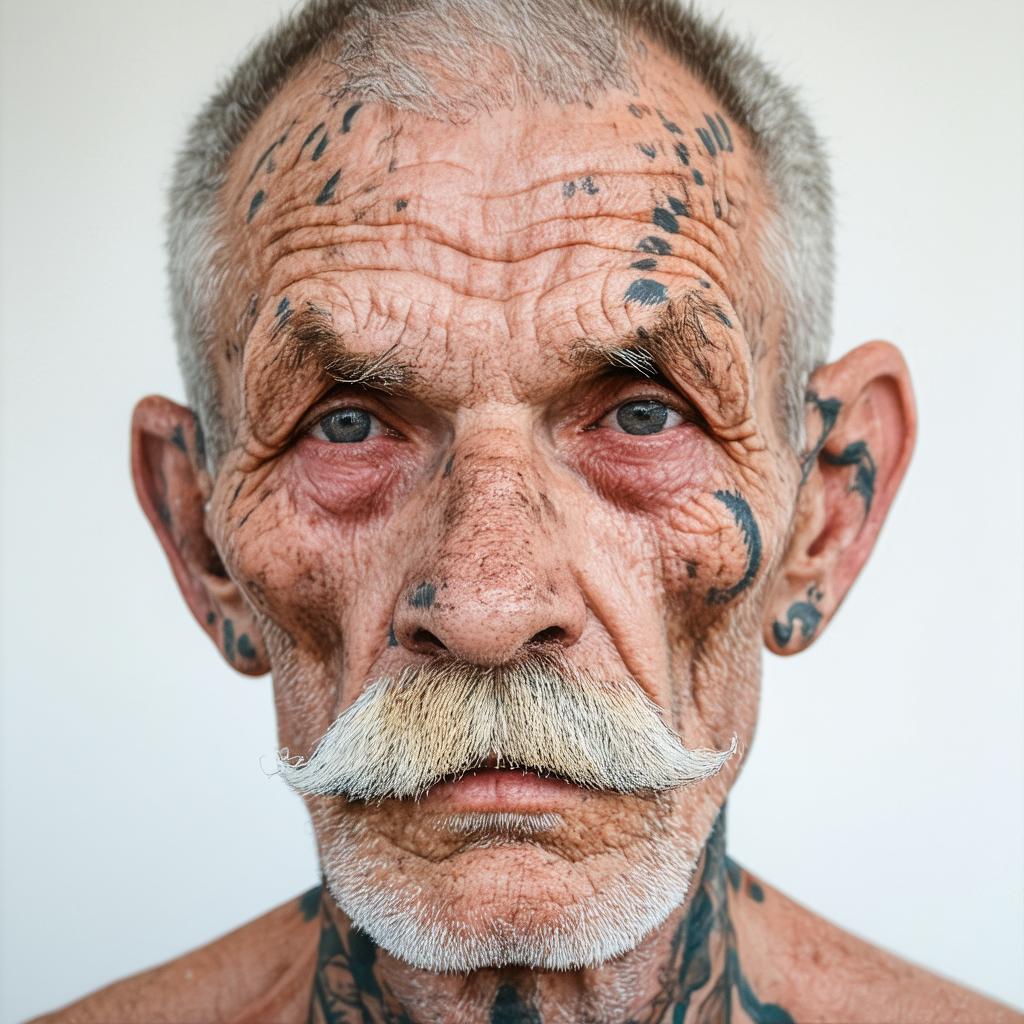}}
\caption[]{Example generated images with the following input prompts (except for universal) and $\mathcal{M}$:
\subref{fig:ex-a} Black androgynous person, \textit{generated by CosmicMan},
\subref{fig:ex-b} East Asian woman, curly hair, having acne scars, normal weight, \textit{generated by SD3.5}, 
\subref{fig:ex-c} Indian boy, wearing color contact lenses, pink hair, slightly chubby, \textit{generated by CosmicMan},
\subref{fig:ex-d} Latino man, \textit{generated by Kandinsky},
\subref{fig:ex-e} Middle Eastern androgynous person, bald, \textit{generated by Flux.1},
\subref{fig:ex-f} Pacific Islander baby boy, auburn hair, \textit{generated by Flux.1},
\subref{fig:ex-g} Pacific Islander girl, bald, over-weighted, having face tattoo, \textit{generated by Kandinsky},
\subref{fig:ex-h} Southeast Asian girl, red hair, \textit{generated by SD3.5} and
\subref{fig:ex-i} White man, having wrinkles, mustache, having face tattoo, \textit{generated by SD3.5}.}
\label{fig:examples}
\end{figure}

Figure~\ref{fig:examples} shows some example images from DIF-V with their corresponding input prompts except for universal ones.

\subsection{Producing Image Variations}

Besides generating images of different identities covering a wide range of realistic face characteristics, we need to create variations of each identity. The goal is to produce multiple variations for each unique identity using different tools. In addition, we want the resulting images to resemble those of actual identity cards.

Therefore, as a first step, we use the HivisionIDPhotos tool \citep{hivision} to remove the background of the images and replace it with a simple basic color. In this way, the images resemble photos suitable for identities or passports. 

\begin{figure}[ht]
\centering
\includegraphics[width=9 cm]{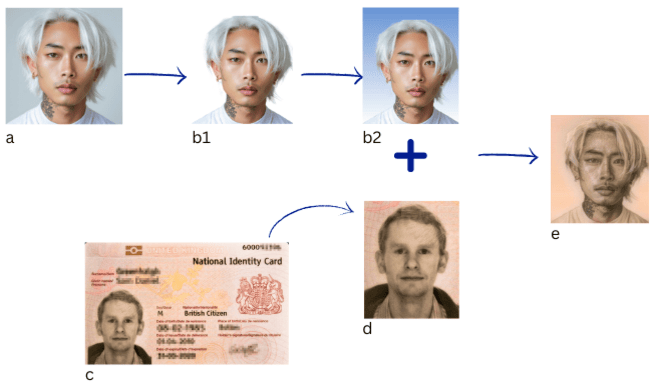}
\caption{Extracting style from identity cards: \textbf{a.} initial face image generated by Stable Diffusion $3.5$, \textbf{b1-b2.} the image after background removal and blue background addition respectively both with HivisionIDPhotos tool, \textbf{c.} the identity card from which face image will be extracted, \textbf{d.} the extracted face image and \textbf{e.} the final image after using the style of d and PAMA tool.}
\label{fig:pama}
\end{figure}

In the next step, we pass the images through certain filters to make them appear as actual identity shots. More precisely, we used $23$ different nations' identity card image templates, derived from the Internet as input to the PAMA tool \citep{luo2022progressive}. The different nations from which we derive identity images are (in alphabetical order): \textit{Australia, Austria, Brazil, Bulgaria, Canada, China, Cyprus, France, Greece, Kenya, Lithuania, Nepal, Netherlands, New Zealand, Pakistan, Poland, Romania, Slovenia, Spain, Turkey, United Kingdom, Ukraine, US (California)}. We extracted the facial image from each of the $23$ identity cards and used it as a style image. In Figure~\ref{fig:pama}, this procedure is shown.

\begin{figure}[ht]
\centering
\includegraphics[width=12.5 cm]{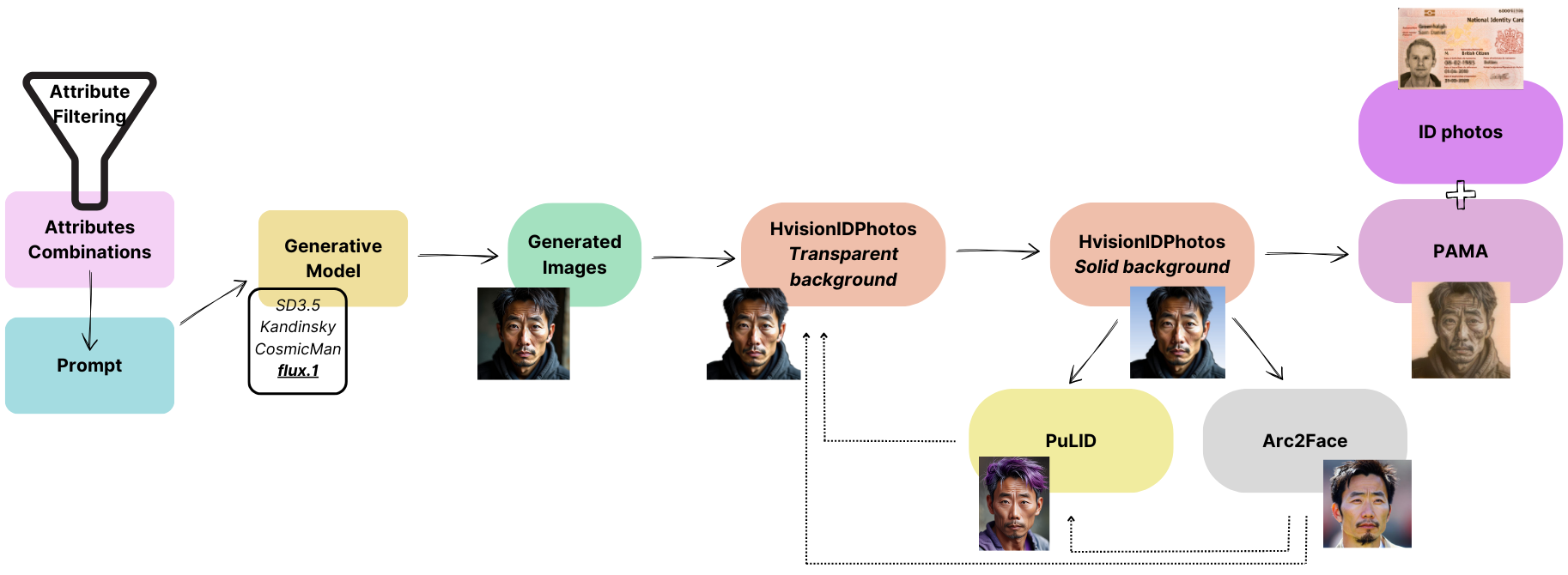}
\caption{General scheme of the proposed framework. After the $\mathcal{A}$ options filtering and combination, the prompt given (except for universal) was \textit{a portrait of an East Asian man, 4K, ultra realistic}. As a following step, the generative model flux.1 was used.}
\label{fig:whole_process}
\end{figure}

To create multiple images for each identity, we utilized two state-of-the-art generative $\mathcal{M}$: Arc2face \citep{papantoniou2024arc2face} and PuLID \citep{guo2024pulid} version v1.1. For Arc2Face, we used a CFG weight of 3, while for PuLID, the CFG weight was set to 5. In both cases, the number of inference steps, $i$, was set to $25$. Additionally, for PuLID, we selected the fidelity mode instead of extreme style to achieve higher prompt adherence. To get even more images for each identity, we used Arc2Face images as input to PuLID tool as shown in Figure~\ref{fig:whole_process}.

Following these changes, we pass the final images from the PAMA and HivisionIDPhotos tools once more, respectively.

\begin{figure}[ht]
\centering
\includegraphics[width=12.5 cm]{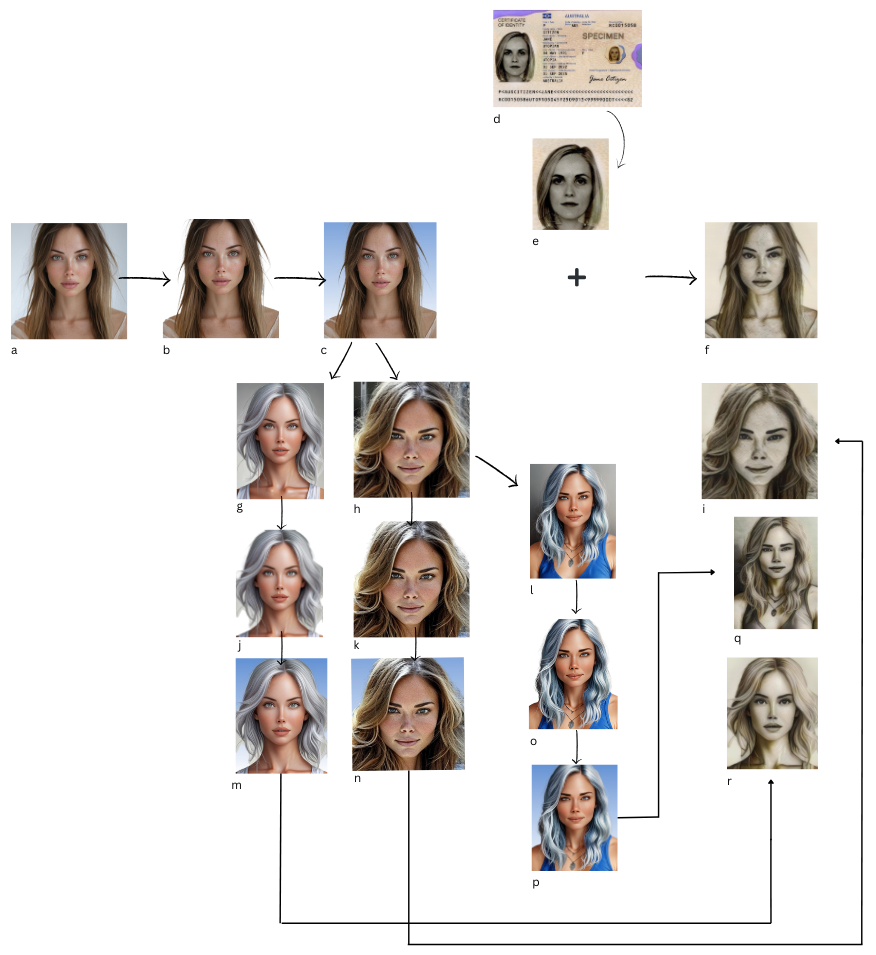}
\caption{\textbf{a.} The initial image generated by Stable Diffusion $3.5$ with the following prompt (except for universal): \textit{a portrait of a White woman, long hair, slightly chubby, Ultra HD, ultra realistic}, \textbf{b.} and \textbf{c.} we removed and applied blue background respectively, utilizing HivisionIDPhotos tool, \textbf{f.} after using the PAMA tool, the image of c. was filtered to look more like the identity image of d. \textbf{g.} to get an alternate of the initial image, we utilized the PuLID tool with the following prompt: \textit{woman, White, portrait photo, front face, realistic, color, detailed face, white hair, hazel eyes}, \textbf{h.} we used the Arc2face tool to get an alternate of the initial image, \textbf{j., m., r., k., n., i.} depict the images of g and h after the use of Hivision and PAMA tools respectively, \textbf{l.} depicts an alternate image of that of h given the prompt: \textit{woman, White, portrait photo, front face, realistic, color, detailed face, blue hair, hazel eyes}, finally \textbf{o., p., q.} depict the image of l after the use of Hivision and PAMA tools respectively.}
\label{fig:framework}
\end{figure}

All the process steps are illustrated in Figure~\ref{fig:whole_process} and an example is shown in Figure~\ref{fig:framework}.

The final dataset includes $27,780$ distinct face images corresponding to $926$ unique identities. We generated approximately $230$ images with each of the four different generative $\mathcal{M}$ as evenly distributed in terms of gender and race as possible: around $300$ images per gender and $115$ identities across eight racial groups. The modified images per identity are $30$.

\section{Evaluation}
\label{sec:evaluation}

To assess the utility of our suggested face image dataset as a face verification benchmark, we aimed to create a challenging yet fair evaluation protocol that reflects real-world demographic diversity. In existing face verification systems, bias towards certain gender and racial groups can result in uneven performance. Therefore, our goal was to ensure that the evaluation process explicitly considers these demographic $\mathcal{A}$ options, helping to highlight both the strengths and weaknesses of face verification models across different groups. Besides, we also focused on evaluating our dataset concerning the identity style applied.
To this end, we followed the steps outlined below.
We first split the dataset into balanced positive and negative pairs, according to gender and race. In both cases, we used cosine similarity to find the pairs. More precisely, for the positive pairs we compared all the images of the same identity and kept the one with the minimum cosine similarity. In parallel, for the negative pairs, we kept the pair among the same gender or race with the maximum similarity. In this way, the dataset will be more challenging to evaluate \citep{wang2019racial, thom2023doppelver}.
The next step was benchmarking our dataset towards gender and race $\mathcal{A}$ options in terms of certain metrics: verification accuracy, True Acceptance Rate (TAR) at Fixed False Acceptance Rate (FAR), Equal Error Rate (EER), and AUC (Area Under the Curve) from ROC. We utilized the DeepFace framework \citep{serengil2020lightface, serengil2024lightface} for face verification, which provides a unified interface for multiple state-of-the-art face recognition models. Within this framework, we used the MTCNN detector \citep{zhang2016joint} to detect, crop and align faces in images, and then either ArcFace \citep{deng2019arcface} or FaceNet \citep{schroff2015facenet} to extract the numerical embeddings from the detected faces. These verification models were selected since they outperform other similar models according to existing benchmarks \citep{serengil2024lightface}.

Table~\ref{tab:gender} shows the results considering gender, while Table~\ref{tab:race} shows the corresponding race results.


\begin{table}[H]
  \centering
  \begin{tabular}{ p{2.2cm}  p{1.4cm}  p{1.2cm} p{1.2cm} p{1.2cm} p{0.8cm} }
    \toprule
    \mcrot{1}{l}{45}{\textbf{Gender}}   & \mcrot{1}{l}{45}{\textbf{Verification model}} & \mcrot{1}{l}{45}{\textbf{Accuracy}} & \mcrot{1}{l}{45}{\textbf{TAR at 1\% FAR}} & \mcrot{1}{l}{45}{\textbf{EER}} & \mcrot{1}{l}{45}{\textbf{AUC}}   \\
    \midrule

Female  & ArcFace  & 90.52\%     & 33.06\% & 9.28\% & 0.95  \\ 
Male  & ArcFace  & 90.81\%     & 74.92\% & 8.86\% & 0.97  \\
Non-binary    & ArcFace  & 89.82\%   & 57.36\% & 10.15\% & 0.96  \\
Female   & Facenet  & 85.98\%   & 48.39\% & 11.81\% & 0.95  \\
Male   & Facenet  & 88.52\% & 49.49\%  & 7.75\% & 0.97   \\ 
Non-binary  & Facenet  & 89.51\%  & 56.46\% & 8.00\% & 0.97 \\
\bottomrule
  \end{tabular}
\caption{Performance of the verification models on the three different genders of DIV-F dataset.}
  \label{tab:gender}
\end{table}


\begin{table}[H]
  \centering
  \begin{tabular}{ p{3cm}  p{1.4cm}  p{1.2cm} p{1.2cm} p{1.2cm} p{0.8cm} }
    \toprule
    \mcrot{1}{l}{45}{\textbf{Race}}   & \mcrot{1}{l}{45}{\textbf{Verification model}} & \mcrot{1}{l}{45}{\textbf{Accuracy}} & \mcrot{1}{l}{45}{\textbf{TAR at 1\% FAR}} & \mcrot{1}{l}{45}{\textbf{EER}} & \mcrot{1}{l}{45}{\textbf{AUC}}   \\
    \midrule

Black  & ArcFace  & 86.42\%     & 44.58\% & 15.19\% & 0.94  \\ 
Black   & Facenet  & 79.01\%   & 30.12\% & 13.92\% & 0.92  \\
East Asian  & ArcFace  & 77.72\%     & 2.20\% & 18.28\% & 0.89  \\
East Asian   & Facenet  & 63.04\% & 10.99\%  & 18.28\% & 0.89   \\ 
Indian    & ArcFace  & 92.92\%   & 84.43\% & 6.78\% & 0.99  \\ 
Indian  & Facenet  & 89.58\%  & 83.61\% & 11.02\% & 0.98 \\
Latino    & ArcFace  & 90.99\%   & 73.87\% & 12.61\% & 0.96  \\ 
Latino  & Facenet  & 89.19\%  & 54.95\% & 10.81\% & 0.97 \\
Middle Eastern    & ArcFace  & 91.37\%   & 29.13\% & 9.57\% & 0.97  \\ 
Middle Eastern  & Facenet  & 90.86\%  & 5.83\% & 8.51\% & 0.97 \\
Pacific Islander    & ArcFace  & 85.48\%   & 53.91\% & 13.33\% & 0.93  \\ 
Pacific Islander  & Facenet  & 79.44\%  & 59.38\% & 10.83\% & 0.94 \\
Southeast Asian    & ArcFace  & 82.79\%   & 50.40\% & 12.61\% & 0.92  \\ 
Southeast Asian  & Facenet  & 69.67\%  & 38.40\% & 11.76\% & 0.93 \\
White    & ArcFace  & 92.69\%   & 51.33\% & 6.60\% & 0.97  \\ 
White  & Facenet  & 89.50\%  & 76.99\% & 6.60\% & 0.97 \\
\bottomrule
  \end{tabular}
\caption{Performance of the verification models on the eight different races of DIV-F dataset.}
  \label{tab:race}
\end{table}


The following Figures~\ref{fig:charts1} and~\ref{fig:charts2} present graphical representations of the performance metrics outlined in Tables~\ref{tab:gender} and~\ref{tab:race}, which evaluate the DIV-F dataset across gender and race subgroups using two face verification models (FaceNet and ArcFace). The depicted metrics include verification accuracy, True Acceptance Rate (TAR) at 1\% False Acceptance Rate (FAR), and Equal Error Rate (EER).

\begin{figure}[ht]
\centering
\subfigure[][]{
\label{fig:g-a}
\includegraphics[width=\textwidth]{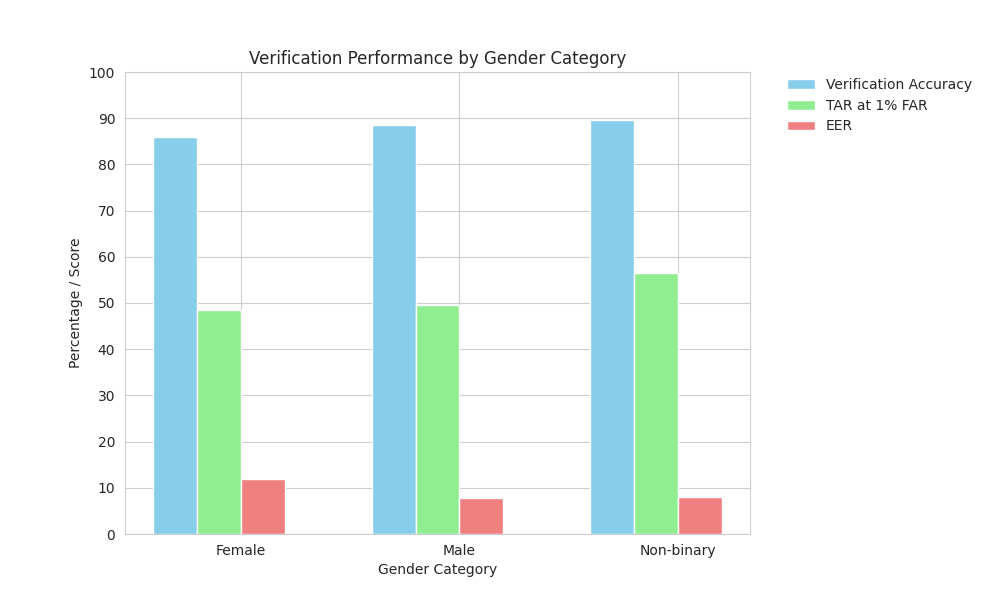}}
\hspace{0.5in}
\subfigure[][]{
\label{fig:g-b}
\includegraphics[width=\textwidth]{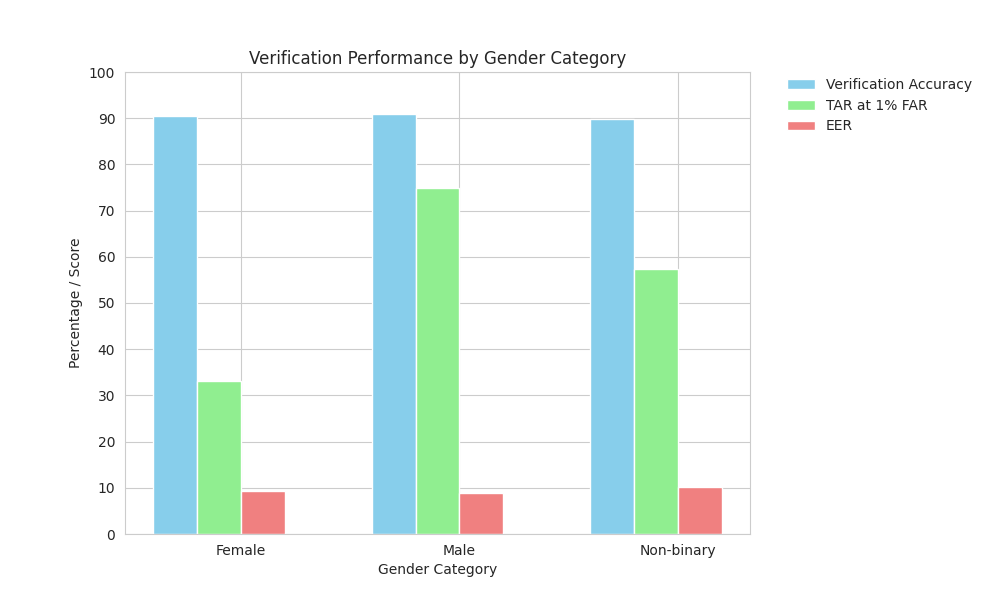}}

\caption[DIV-F performance evaluation]{Performance of the verification models on gender $\mathcal{A}$ options:
\subref{fig:g-a} using ArcFace,  
\subref{fig:g-b} using Facenet}

\label{fig:charts1}
\end{figure}


\begin{figure}[ht]
\centering

\subfigure[][]{
\label{fig:r-a}
\includegraphics[width=\textwidth]{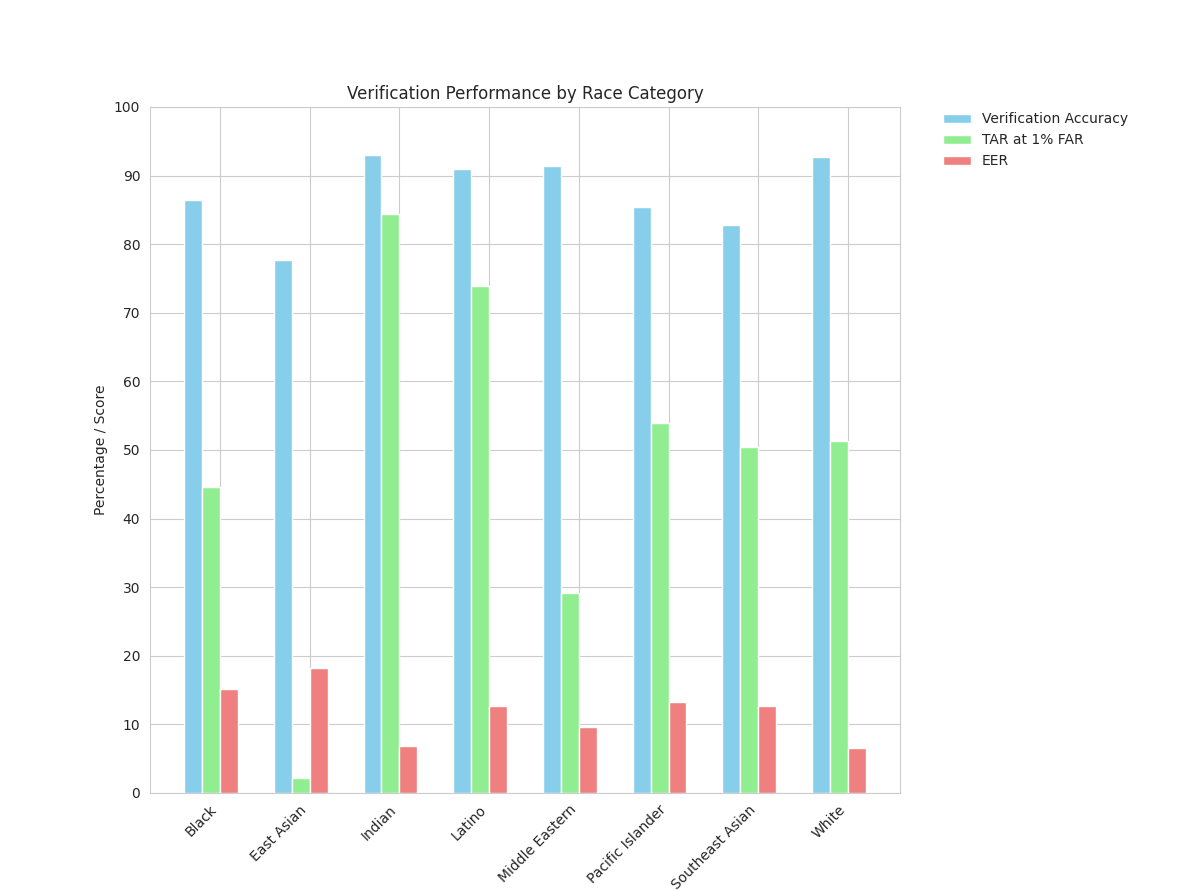}}
\hspace{0.5in}
\subfigure[][]{
\label{fig:r-b}
\includegraphics[width=\textwidth]{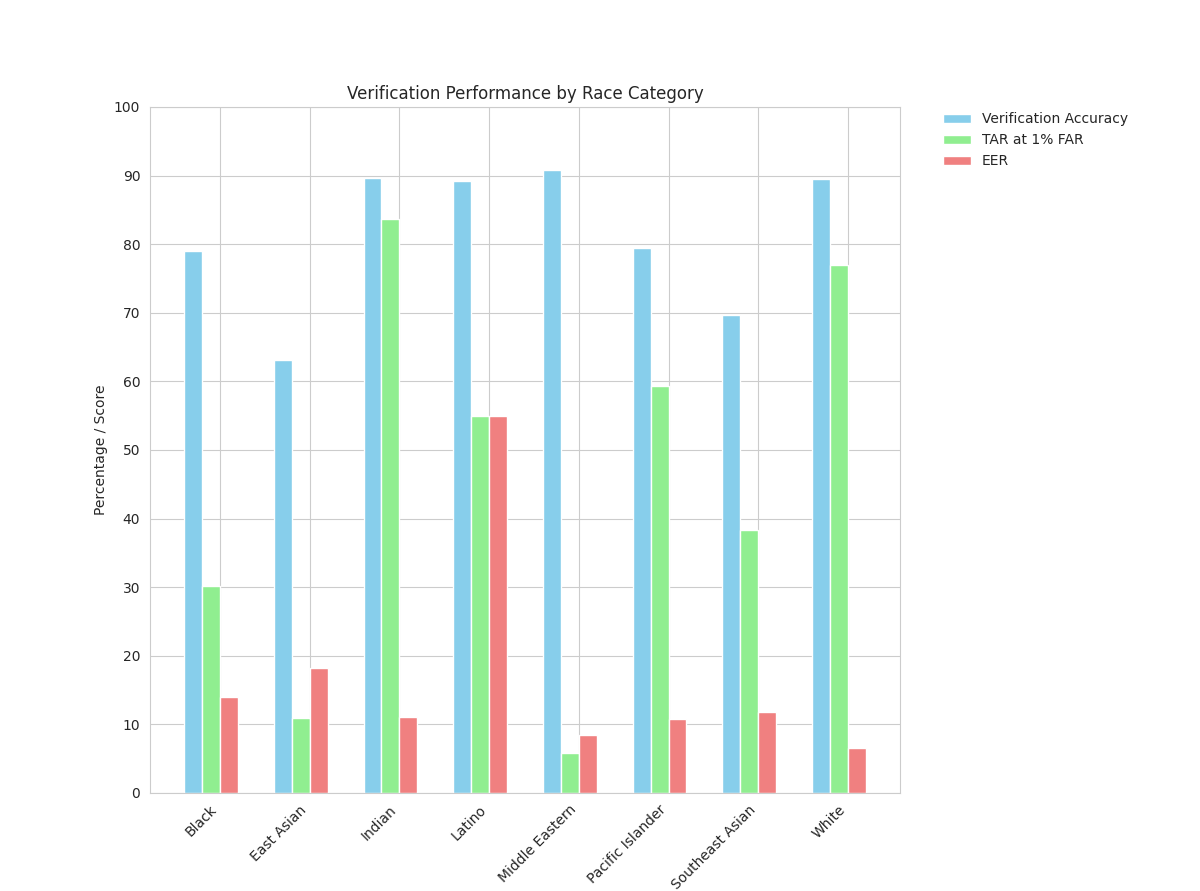}}
\caption[DIV-F performance evaluation]{Performance of the verification models on race $\mathcal{A}$ options:
\subref{fig:r-a} using ArcFace,  
\subref{fig:r-b} using Facenet}
\label{fig:charts2}
\end{figure}


The results indicate that face verification models consistently under perform on Female faces compared to Male and Non-binary faces across the evaluated metrics. This holds true for both ArcFace and Facenet. Furthermore, in gender-specific evaluation, Male faces achieve superior performance, especially when paired with the ArcFace model.
In the race-based evaluation, Asian faces—East Asian and Southeast Asian subgroups—exhibit the lowest performance across all metrics for both the ArcFace and Facenet verification models. In contrast, White and Indian faces demonstrate the best performance, with the most pronounced performance observed when using the ArcFace model.

In the following, we examine the impact of identity (ID) style on the performance of face verification models. Specifically, we compare the performance of ArcFace and Facenet on the DIF-V dataset both before and after applying ID style modifications. Evaluation metrics include verification accuracy, True Acceptance Rate (TAR) at 1\% False Acceptance Rate (FAR), Equal Error Rate (EER), and Area Under the Curve (AUC).
Table~\ref{tab:id_perf} shows the corresponding results.


\begin{table}[H]
  \centering
  \begin{tabular}{ p{1.4cm}  p{1.4cm}  p{1.2cm} p{1.2cm} p{1.2cm} p{0.8cm} }
    \toprule
    \mcrot{1}{l}{45}{\textbf{ID style}}   & \mcrot{1}{l}{45}{\textbf{Verification model}} & \mcrot{1}{l}{45}{\textbf{Accuracy}} & \mcrot{1}{l}{45}{\textbf{TAR at 1\% FAR}} & \mcrot{1}{l}{45}{\textbf{EER}} & \mcrot{1}{l}{45}{\textbf{AUC}}   \\
    \midrule

\tikzxmark   & ArcFace  & 94.67\%     & 82.07\% & 5.32\% & 0.99  \\ 
\tikzxmark  & Facenet  & 90.34\%   & 82.41\% & 6.35\% & 0.99  \\
  \midrule
\checkmark  & ArcFace  & 91.61\%     & 57.99\% & 8.44\% & 0.97  \\
\checkmark   & Facenet  & 90.24\% & 39.48\%  & 7.84\% & 0.97   \\ 
\bottomrule
  \end{tabular}
\caption{Performance of the verification models before and after applying ID style on images of DIV-F dataset.}
  \label{tab:id_perf}
\end{table}

\begin{figure}[ht]
\centering
\includegraphics[width=15cm]{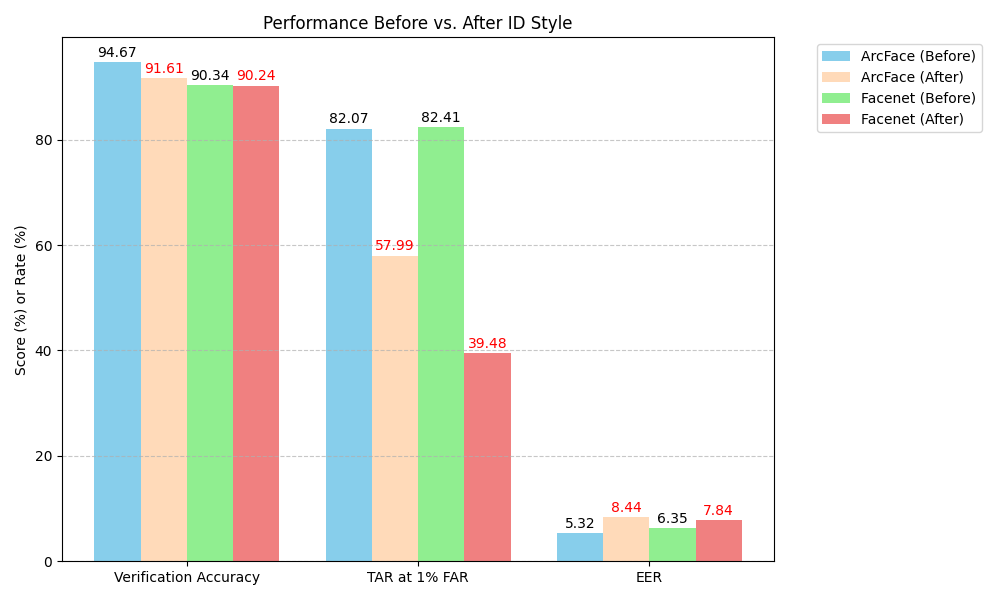}
\caption{Performance of the verification models before and after applying ID style on DIF-V images.}
\label{fig:charts3}
\end{figure}

Figure~\ref{fig:charts3} presents graphical representations of the performance metrics outlined in Table~\ref{tab:id_perf}. The depicted metrics include verification accuracy, TAR at 1\% FAR, and EER.


The results demonstrate that applying identity (ID) style to facial images degrades the performance of both the ArcFace and Facenet verification models. This suggests that ID style transformations introduce additional challenges to the face verification task for these models.

\section{Ethical Issues and Risks}
\label{sec:ethical}

Generally, text-to-image generation is a primarily data-driven process. Therefore, generative models trained on massive amounts of unfiltered data might suffer from, or even amplify, data biases that pose ethical concerns \citep{chen2025comprehensive}. This issue is particularly evident in scenarios that involve gender and/or race. Furthermore, the main source of this problem seems to be that these generative models rely on publicly available Internet data, which inevitably carry the biases and prejudices that are present in society. Unfiltered data may contain implicit social prejudices, preconceptions, and potentially offensive or dangerous content \citep{hao2024synthetic}.

In addition, since model-generated images are as realistic as actual photographs, they can be utilized to achieve misleading and misleading objectives, affecting public opinion and the veracity of information \citep{hu2023loss, hong2024all}. Quick dissemination of such content via social media platforms exacerbates the problem, allowing misinformation to go viral and making it extremely difficult to contain or review. Furthermore, identifying the source of AI-driven misinformation poses a major issue. AI-generated content frequently remains anonymous or is wrongly attributed, making it difficult to track creators \citep{kumar2024role}.


\section{Conclusions}
\label{sec:conclusions}

The current work highlights how important it is for face image datasets to be equitable and diverse in the context of face verification tasks. Due to the prevalence of gender, racial, or other demographic biases in current real-world datasets, verification systems may become less reliable and equitable. By leveraging advanced generative models, we have developed a comprehensive methodology that enables the creation of synthetic face image datasets that more accurately represent the demographic diversity of the real world. The introduction of the Diverse and Inclusive Faces for Verification (DIF-V) dataset offers a significant contribution to this challenge, providing a strong resource for practitioners and researchers looking to reduce biases in current face verification techniques. Comprised of $27,780$ images from $926$ unique identities, the DIF-V dataset not only serves as a first-to-know synthetic benchmark for testing and evaluating face verification systems, but also emphasizes the importance of including diverse representations in AI-based technologies. This work lays the foundation for future developments that aim to improve the inclusivity and effectiveness of face verification technologies as the need for accuracy and fairness in biometric systems grows only. Our strategy advances efforts to create systems that are not only efficient, but also equitable and representative of all people by tackling the inherent inequities of existing datasets. It also adds to the continuing discussion about ethical AI practices.

\section{Acknowledgments}

\label{sec:acknowledgments}
This research work was funded by the European Union under the Horizon Europe projects MAMMOth (Grant Agreement ID: 101070285) and ELIAS (Grant agreement ID: 101120237).

\bibliographystyle{elsarticle-harv} 
\bibliography{references}

\end{document}